\RequirePackage{fix-cm}
\documentclass{svjour3}                     
\smartqed  
\usepackage{graphicx}
\usepackage{rotating}
\usepackage[utf8]{inputenc} 
\usepackage[T1]{fontenc}    
\usepackage{url}            
\usepackage{booktabs}       
\usepackage{amsfonts}       
\usepackage{nicefrac}       
\usepackage{microtype}      
\usepackage{lscape}

\usepackage[nofillcomment,ruled,linesnumbered]{algorithm2e}
\usepackage{amsmath}
\usepackage{xcolor}
\usepackage{graphicx}

\usepackage{enumitem}

\definecolor{orange}{rgb}{1.0, 0.5, 0.0}

\newcommand{\bbs}{\mbox{$\mathbb{S}$}}

\begin{document}

\title{Explainable Clustering via 
Exemplars: Complexity and Efficient Approximation Algorithms}
\titlerunning{Explainable Clustering via Exemplars}


\author{Ian Davidson \and
        Michael Livanos \and
        \newline Antoine Gourru \and
        Peter Walker \and
        \newline Julien Velcin \and
        S. S. Ravi
}
\institute{
  I. Davidson \at 
      Computer Science Department,
     University of California Davis, USA\\
     \email{davidson@cs.ucdavis.edu} \and
  M. Livanos \at 
      Computer Science Department,
     University of California Davis, USA\\
     \email{mjlivanos@ucdavis.edu} \and
  A. Gourru \at 
      Department of Mathematics and Informatics,
     University of Lyon, France\\
     \email{antoine.gourru@gmail.com} \and
  P. Walker \at
     Joint AI Center (JAIC), USA \\
     \email{peter.b.walker.mil@mail.mil } \and
  J. Velcin \at 
      Department of Mathematics and Informatics,
     University of Lyon ERIC LYON 2, France\\
     \email{Julien.Velcin@univ-lyon2.fr} \and
  S. S. Ravi \at
     Biocomplexity Institute, University of Virginia and \\
     Department of Computer Science, University at Albany -- State University
     of New York, USA \\ 
     \email{ssravi0@gmail.com}
}

\date{Received: date / Accepted: date}


\newtheorem{fact}{Fact}[section]
\newtheorem{observation}{Observation}[section]

\newcommand{\cnp}{\textbf{NP}}
\newcommand{\true}{\texttt{True}}
\newcommand{\false}{\texttt{False}}

\newcommand{\QED}{\hfill\rule{2mm}{2mm}}

\newcommand{\irange}{\mbox{$1 \leq i \leq n$}}
\newcommand{\jrange}{\mbox{$1 \leq j \leq m$}}

\newcommand{\dunder}[1]{\underline{\underline{#1}}}



\newcommand{\cscce}{\mbox{SCCE-CE}}

\newcommand{\cale}{\mbox{$\mathcal{E}$}}
\newcommand{\opt}{\mbox{OPT}}

\newcommand{\ian}[1]{\textcolor{red}{#1}}
            
\maketitle

\begin{abstract}
Explainable AI (XAI) is an important developing area but 
remains relatively understudied
for clustering. We propose an explainable-by-design clustering
approach that not only finds clusters but also exemplars to explain
each cluster. The use of exemplars for understanding is supported by the 
exemplar-based school of concept definition in psychology. 
We show that finding
a small set of exemplars to explain even a single cluster 
is computationally intractable; hence,
the overall problem is challenging. We develop an approximation
algorithm that provides provable performance guarantees with
respect to clustering quality as well as the number of exemplars
used. This  basic algorithm explains all the instances in 
every cluster whilst another approximation algorithm 
uses a bounded number of exemplars to allow simpler
explanations and provably covers a large fraction of all the instances.
Experimental results show that our work is useful in domains
involving difficult to understand deep embeddings of images and text.
\end{abstract}

\medskip


\section{Introduction}
\label{sec:intro}



The area of explainable AI (XAI) tries to make the complex results
of an algorithm interpretable by humans. Most work has focused on
supervised learning \cite{gunning2019darpa}, and in particular,
\emph{instance-level} explanations such as which parts of an image
resulted in a certain prediction \cite{ribeiro2016should}. Our work
differs from most XAI work in several ways.  Firstly, we explore
unsupervised learning, and in particular, clustering. Secondly, we
seek higher level explanations of the \emph{entire} clustering and
not just why an instance was placed in a particular cluster. This
is not only an understudied problem, but one where explanation is
most needed due to the lack of ground truth annotations (i.e.,
classes) around which explanations can be built.

Existing work in explainable clustering generate explanations in
terms of the \underline{underlying features} used in the clustering
\cite{frost2020exkmc,michalski1983learning,saisubramanian2020balancing}.
These methods are not suitable for modern settings that use
non-interpretable features such as those produced by auto-encoders,
word embeddings (e.g., BERT \cite{devlin2019bert}) 
or graph embeddings (e.g., \cite{Grover-etal-2016}). Consider two
settings which we use to demonstrate our work: clustering of deep
embeddings of sentences and images. In this context, the dimensions
of the embedded space are completely meaningless to a human. Even
if they are eventually understood, the number of dimensions (often
hundreds or more) used by deep embedding poses significant challenges.

\medskip

\noindent
\textbf{Core Idea.} We address the need for explanation in complex
data by creating an exemplar-based approach to clustering that
simultaneously finds clusters of points and exemplars that characterize
the clusters.  We say that an instance $x$ explains another instance
$y$ (or instance $x$ serves as an \textbf{exemplar} for instance $y$)
if $y$ falls within $\epsilon$ distance of $x$ (i.e., $y$ is
within the ball of radius $\epsilon$ centered at $x$). 

Exemplars are a natural mechanism for explanation of concepts
\cite{murphy2004big} by enumerating the  different variations of
the concept.  Consider the situation of explaining a cluster of
images of a single person, say former French President, Jacques
Chirac (center images in Figure \ref{fig:faceExemplars}). One could
describe him as `balding', `tall' and with a `cherubic face'. But
this requires such information to be available for each and every
image; moreover, such a description only fits him when he was the
President of France after he turned 60.  When he was the Prime
Minster of France (during his 40's) he looked very different.  Hence,
an alternative explanation of ``What does Jacques Chirac look like?"
is through exemplars of what he looked like over the years and in
different poses. The purpose of this work is to choose such exemplars
and discover clusters.  Cognitive science literature
(e.g., \cite{walsh2010multilevel}) indicates that exemplars are ideal for explaining
complex concepts/clusters.  Simply increasing $k$ and using 
the resultant centroids or finding sub-clusters 
within clusters \cite{KMEAP-Wang-Chen-2014}  does not address 
this challenge as in many situations there is 
a natural number of clusters (e.g., Figure \ref{fig:mnist}).  
Further,  the variations of the concept need not find dense 
sub-clusters as shown in Figure \ref{fig:mnist}.


\medskip

\noindent
\textbf{Contributions.} Our contributions are as follows.

\begin{enumerate}
\item We formulate the novel explainable clustering  via 
exemplars problem\footnote{The differences between our approach and
density-based clustering (e.g., DBSCAN \cite{Ester-etal-1996}) and multi-centroid clustering (e.g. \cite{KMEAP-Wang-Chen-2014})
are covered in the related work section.} 
and show that even explaining a single cluster 
is a computationally intractable problem (Theorem~\ref{pro:mse_hardness}).

\item Our setting is naturally a bi-objective clustering problem
with respect to cluster quality and explanation quality but we
simplify parameter choice by binding both objectives together with
the same  parameter $\epsilon$.

\item We propose a polynomial time clustering algorithm 
(Algorithm \ref{alg:scce_approx}) that provides provable
performance guarantees with respect to both the maximum cluster diameter  and
the minimum number of exemplars.
More precisely, the maximum cluster diameter is
$2(D^*+\epsilon)$, where $D^*$ is the optimal diameter whilst using at most
$O(N^*\log{n})$ exemplars, where $N^*$ is the  
minimum number of exemplars needed for the dataset of size $n$ (Theorem~\ref{thm:perf_ssce_approx}).  

\item We also provide a relaxed version of the  algorithm 
(see Algorithm \ref{alg:sccre_approx}) that 
upper bounds the number of exemplars by relaxing the requirement to
explain every instance in the cluster.
This algorithm provides the same performance guarantee with
respect to maximum diameter as our previous algorithm. 
The number of instances covered by exemplars is 
at least $(1-1/\mathrm{e}) Q^*$ (which is $\approx 0.63\,Q^*$),
where $\mathrm{e}$ is the base of the natural logarithm and $Q^*$ is the maximum
number of instances that can be covered, given the bound on the number
of exemplars (Theorem~\ref{thm:perf_sscre_approx}).  

\item The algorithms mentioned in Items 3 and 4 
are obtained by combining classic approximation
algorithms from the literature which are implemented in a variety of packages and platforms. This allows for ease of implementation and scalability (possibly via parallelism) and in the repository of our work we provide Python implementations using standard packages.
A novelty of our contribution is that such a combination provides
provable worst-case performance guarantees and
also shows very good experimental performance.

\item We experimentally evaluate our methods on several domains
involving deep embeddings of images (Faces in the wild), text (a Harry Potter novel) and on MNIST digits. We
also begin to explore the novel direction of using exemplars for
another ML task, namely transfer learning.  
\end{enumerate}

\noindent
\textbf{Organization.}~
We begin with an overview of our method and then our problem
definitions.
This is followed by our complexity result and approximation
algorithms. Then, we discuss our experimental
results and finally conclude.\newline



\section{Overview of Our Approach}

\begin{figure}
\centering
\includegraphics[scale=0.45]{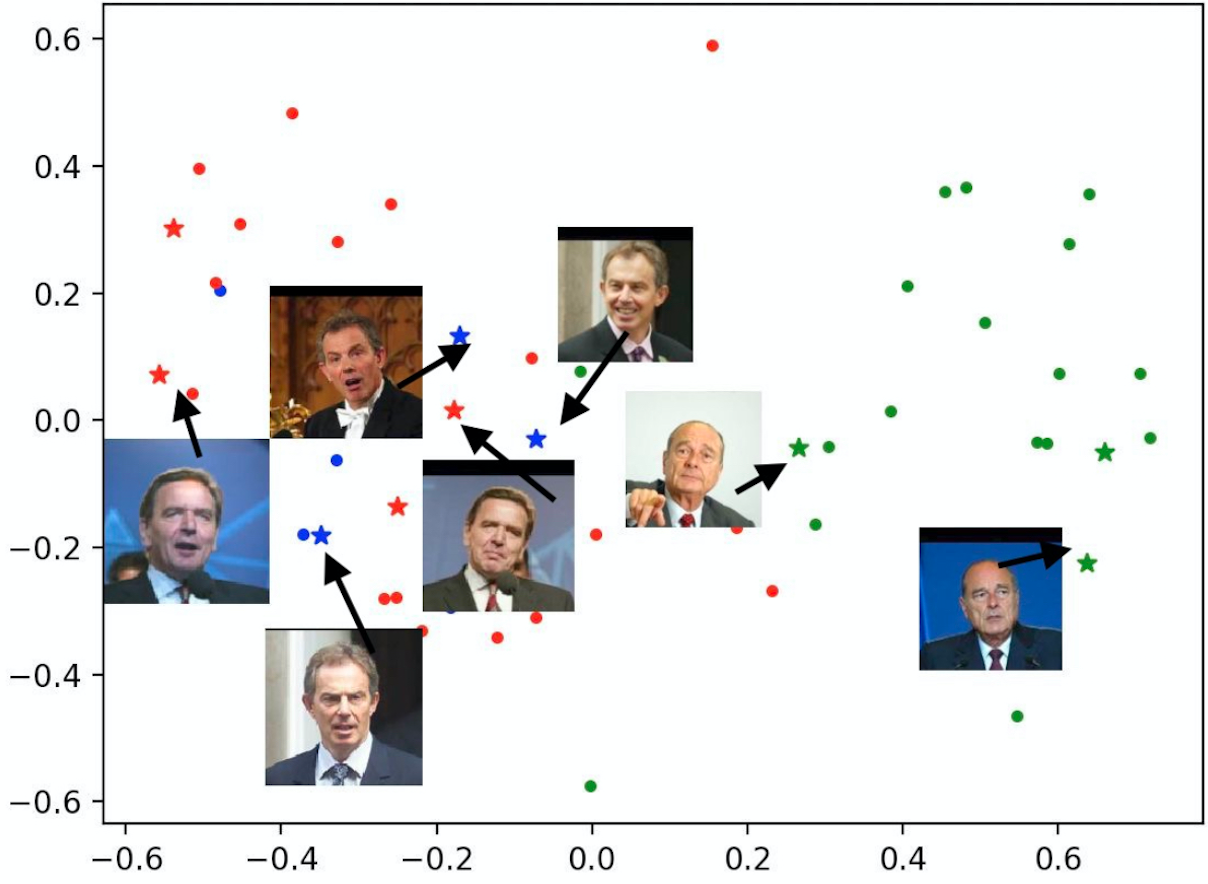}
\caption{An illustrative example of generating clusters (color) and
selecting exemplars (stars). The exemplars form a prototypical
explanation of a cluster in that they cover all instances in the
cluster.  Note the exemplars need not be (and rarely) are close to the centroids.} 
\label{fig:example_alg_output}
\end{figure}

The input to our method is a collection of instances that we wish
to both cluster and explain. Hence, our method is an example of an
explainable-by-design clustering algorithm, unlike our previous work that attempts
to find an explanation for a given clustering 
\cite{davidson2018cluster}.
Further, unlike prior work on conceptual clustering,  we do not use the
features used to cluster in developing an explanation; for instance, 
the work discussed in \cite{frost2020exkmc} simultaneously builds a clustering
and a decision tree using the same features. Here, we instead find
a clustering and a suitable subset of the instances (which we call
exemplars) within each cluster to explain it. We say an exemplar
explains a set of instances that are within $\epsilon$ distance of it.
In practice, exemplars are significantly different from cluster
centroids; see Figure \ref{fig:example_alg_output} for an example.

\medskip

\noindent
\textbf{Trading Off Explanation Complexity Against Clustering Quality.}
We design clustering algorithms
that ensure that the maximum diameter of the clustering found is
within a small constant factor of the optimal diameter and 
$\epsilon$ (the radius of an exemplar's coverage).
Hence, the parameter $\epsilon$ provides a natural way to trade off explanation
complexity against cluster compactness. If we make $\epsilon$ small,
we naturally will require more exemplars but will find more compact
clusters. Conversely, if we make $\epsilon$ large, we will create
simpler explanations but at the cost of a larger cluster diameter.
We present efficient approximation algorithms that provide
provable performance guarantees with respect to both the maximum
diameter and the number of exemplars used.  

\medskip

\noindent
\textbf{Exemplars for Explanation and Their Benefits.} Our work can be considered
as a quantification of the exemplar-based school of concepts
\cite{murphy2004big} as we are discovering concepts (the clusters)
and the exemplars that typify/explain them. This contrasts with a 
feature based explanation
(e.g., using the attributes/properties of the face) as described above.
In this paper, we argue that using exemplars has
\underline{pragmatic} and \underline{pedagogical} benefits. As ML/DM
progresses to more complex representations of complex objects, 
using features as the basis for explanation is no longer always valid, even
though there is excellent work in this area \cite{frost2020exkmc}. In
settings where features are not interpretable 
(e.g., deep embeddings of image data), one
pragmatic explanation mechanism is exemplars.
The pedagogical benefit stems from cognitive psychology's 
experimentally-verified rich literature on how humans
understand and comprehend the world; this literature
comes under a topic known as Concept Theory \cite{murphy2004big,ashby2005human}. 
In particular, exemplars are a natural explanation vehicle as they
leverage the existing knowledge of humans to make internalizing the
explanation easier. For example, the exemplars of Jacques Chirac
will be internalized differently by say a French citizen 
(e.g., ``he looks
like a taller version of former French President Fran\c{c}ois Mitterrand'')
versus an American citizen (e.g., ``he is as tall as and looks like the former US
President Ronald Reagan''). This direction also presents the opportunity
to exploit the literature on how humans organize exemplars into
ontologies or hierarchies for more complex explanations 
\cite{walsh2010multilevel}.

\medskip

\noindent
\textbf{Difficulty of the Problem.} Our computational problem 
inherently has two intertwined tasks: (i)
finding compact clusters and (ii) finding a minimal set of exemplars to
represent each cluster. This is a challenging problem as 
the first problem is known to be \cnp-hard \cite{Gon-1985} and
additionally we show that even for a single cluster, finding a
minimal set of exemplars to represent the cluster is \cnp-hard 
(Theorem~\ref{pro:mse_hardness}). 
Solving these tasks separately could
yield sub-optimal results; instead, we bind them together using	
a single parameter $\epsilon$ (the exemplar coverage distance)
to simultaneously perform clustering and exemplar selection.
Our algorithms provide provable performance guarantees.

\section{Definitions}
\label{sec:def_prelim}

\subsection{Basic Definitions}\label{sse:basic_defs}

\noindent
Let $X = \{x_1, x_2, \ldots, x_n\}$ be a set of $n$ instances.
We assume that for each pair of instances $x_i$ and $x_j$, we have
a (symmetric) distance $d(x_i, x_j)$. 
The distance function $d$ is assumed to be a \emph{metric}; 
it may be the distance in some embedding space.
We are also given a value $\epsilon > 0$ which is set 
by a domain expert and naturally trades off explanation complexity 
against cluster compactness. 

\smallskip
\noindent
\textbf{Notion of Explanation.} 
Given two instances $x_i$ and $x_j$, where $d(x_i, x_j) \leq \epsilon$,
we say that $x_i$  covers $x_j$ and that $x_i$ 
is an \underline{exemplar} for $x_j$.
Since the distance function $d$ is symmetric, it is also true
in this case that $x_j$ covers $x_i$ and $x_j$ is an exemplar for $x_i$.
For convenience, we will also say that $x_i$ is an $\epsilon$\emph{-neighbor} 
of $x_j$ (and vice versa).
We now generalize this definition to clusters.
Given a subset $Y \subseteq X$ of instances and 
another subset $\cale{} \subseteq X$ of exemplars, 
we say that \cale{} \underline{covers} $Y$ if for every instance $x_i \in Y$,
there is an instance $x_j \in \cale$ such that $x_j$ covers $x_i$
(i.e., $x_j$ is an exemplar for $x_i$). 
When a subset $\cale{} \subseteq X$ of exemplars 
covers a set $Y \subseteq X$, we say that \cale{} 
is an \underline{exemplar set} for $Y$ and that 
$Y \cup \cale{}$ forms a cluster and its explanation.

For any instance $x_i$, let $S_i \subseteq X$ consist of all the instances
that are $\epsilon$ neighbors of $x_i$;
that is, $d(x_i, x_j) \leq \epsilon$ for each $x_j \in S_i$.
We refer to $S_i$ as the $\epsilon$-\emph{neighborhood} of $x_i$.
Note that each subset $S_i$ is nonempty since it includes $x_i$ itself.
Further, $x_i$ is an exemplar for all the instances in $S_i$.

\smallskip
\noindent
\textbf{Clustering to Minimize the Maximum Diameter.}~
For clustering a set $X$ of instances, a common objective
is to minimize the maximum diameter \cite{Gon-1985}.  
For the reader's convenience, we provide the associated definitions.
The diameter of any cluster is the maximum
distance between any pair of instances in that cluster.
The diameter of a clustering is the largest
cluster diameter.
It is known that finding a clustering with $k \geq 3$ clusters
that minimizes the maximum diameter is \cnp-hard \cite{GJ-1979}.
When the distance function is a metric, a well known approximation
algorithm due to Gonzalez \cite{Gon-1985} provides a clustering whose
maximum diameter is at most twice the optimal diameter. 

\subsection{Additional Definitions}\label{sse:addl_def}

\noindent
\textbf{Graph Theoretic Definitions:}~
We use some graph theoretic concepts and a special class of
graphs in proving our results.
Given an undirected graph $G(V,E)$, a subset $V'$ of nodes
forms a \textbf{dominating set} for $G$ if for every node $w \in V-V'$,
there is a node $v \in V'$ such that the edge $\{v,w\}$ is in $E$.
Given a graph $G(V,E)$, the goal of the \textbf{minimum dominating set} (MDS) 
problem is to find a dominating set of minimum cardinality for $G$. 

Given a set of disks (i.e., circles in two-dimensional space)
each with the same radius $r$,  
one can define an associated undirected graph
as follows: there is
one node for each disk; there is an edge between two nodes if
the corresponding disks touch or intersect (i.e., the distance between
the centers of the disks is at most $2r$).
Such a graph is called a \textbf{unit disk graph} \cite{CCJ-1990}.
Many optimization problems, including
the MDS problem, are known to be \cnp-hard
even for unit disk graphs \cite{CCJ-1990,HM-1985}.
We rely on the \cnp-hardness of the MDS problem for unit disk graphs in proving
Theorem~\ref{pro:mse_hardness}. 

Unit disk graphs can be defined in three or more dimensions where  each object is a ball of unit radius 
in an appropriate dimension.
Each node of the corresponding graph represents a ball with
an edge between two nodes if their corresponding balls touch/intersect.

\medskip

\noindent
\textbf{Minimum Set Cover (MSC) Problem:}~
In this problem \cite{GJ-1979},
the input consists of a base set $U = \{u_1, u_2, \ldots, u_n\}$,
a collection $Y = \{Y_1, Y_2, \ldots, Y_m\}$, where each $Y_j$ is a subset
of $U$ ($1 \leq j \leq m$) and an integer bound $\beta \leq m$.
The goal is to choose a subcollection $Y'$ of $Y$ with $|Y'| \leq \beta$
such that the union of the sets in $Y'$ is equal to $U$ (i.e., the 
union covers all the elements in $U$).
This problem is \cnp-complete and a natural greedy approximation
algorithm (which picks a new set in each iteration such that the set
covers as many new elements as possible) is known to give a
performance guarantee of $O(\log{n})$ for the problem \cite{Vaz-2001}.
One of our results (Section~\ref{sse:sccre_approx}) 
uses this approximation algorithm.

\medskip

\noindent
\textbf{Budgeted Maximum Coverage Problem:}~
We also use a known approximation algorithm for the 
Budgeted Maximum Coverage (BMC) problem, which is closely related to
the Minimum Set Cover (MSC) problem \cite{GJ-1979}.
The input to the BMC problem is a base set $U = \{u_1, u_2, \ldots, u_n\}$,
a collection $Y = \{Y_1, Y_2, \ldots, Y_m\}$, where each $Y_j$ is a subset
of $U$ ($1 \leq j \leq m$) and a budget $\beta \leq m$.
The goal is to choose a subcollection $Y'$ of $Y$ with $|Y'| = \beta$
such that the union of the sets in $Y'$ covers the maximum
number of elements of $U$.
This problem is also \cnp-hard and a natural greedy approximation
algorithm (which picks a new set in each iteration such that the set
covers as many new elements as possible) has been shown to give a
performance guarantee of $(1-1/\mathrm{e})$ for the problem \cite{Khuller-etal-1999},
with $\mathrm{e}$ being the base of the natural logarithm.
One of our results (Section~\ref{sse:sccre_approx}) uses this result.

\subsection{Main Problem Formulations}
\label{sse:main_formulation}

We now provide rigorous formulations of the 
problems considered in this paper.
We begin with the problem of finding a small set of
exemplars for a given set of instances.

\medskip

\newcommand{\msec}{\mbox{MSEC}}

\noindent
(a) \textbf{Minimum Set of Exemplars for a Cluster} (\msec)

\smallskip
\noindent
\underline{\textsf{Given:}}~ A cluster
$X = \{x_1, x_2, \ldots, x_n\}$ of $n$ instances,
a value $\epsilon > 0$, an integer $\beta \leq |X|$. 

\smallskip
\noindent
\underline{\textsf{Question:}}~ Is there a subset $\cale{} \subseteq X$,
with $|\cale| \leq \beta$, such that \cale{} is an exemplar set for $X$?

We note that the \msec{} problem requires an exemplar set for 
\emph{all} the instances in the set $X$.
We now develop formulations where the set $X$ must be partitioned into
clusters and exemplar sets must be found for each cluster. 
We first provide a formulation where each instance must have an exemplar.

\medskip

\newcommand{\scce}{\mbox{SCCE}}

\noindent
(b) \textbf{Simultaneous Construction of Clusters and Exemplars} (\scce)

\smallskip
\noindent
\underline{\textsf{Given:}}~ 
A set $X = \{x_1, x_2, \ldots, x_n\}$ of $n$ instances to be clustered,
integer $k$, where $2 \leq k \leq n$ (the number of clusters), and 
a value $\epsilon > 0$.

\smallskip
\noindent
\underline{\textsf{Requirement:}}~ Find a partition of $X$ 
into $k$
clusters $C_1$, $C_2$, $\ldots$, $C_k$ and an exemplar set $\cale_j$ 
for each cluster $C_j$, $1 \leq j \leq k$, such that all the following 
conditions hold:
\begin{itemize}
\item
\underline{Compactness of Clustering and Explanation:}~
(i) the maximum diameter of the clusters is as small as possible,
(ii) $\sum_{j=1}^{k}|\cale_j|$ (i.e., the total number of
exemplars used) is as small as possible.
\item
\underline{Distinctness of Explanations:}~
(iii) $\cale_a \cap \cale_b = \emptyset$ for all $1 \leq a,b \leq k$
and $a \neq b$ (i.e., the exemplar sets are pairwise disjoint), and
\item
\underline{Completeness of Explanations:}~
(iv) for each instance $x \in X$, there is an exemplar $y$ such
that $x$ and $y$ are in the same cluster.
\end{itemize}

\smallskip

We will present an approximation algorithm for \scce{} in 
Section~\ref{sec:main_prob_analysis}.  However, this solution may use a large
number of exemplars due to the completeness requirement.
This can make it difficult for a user
to interpret the explanation. To address this, we next explore a relaxed
version of the problem where not all instances
are explained. (Our approximation algorithm for this
problem allows us to analytically bound the number of instances
that are not explained.) 


\newcommand{\sccre}{\mbox{SCCRB}}

\medskip

\noindent
(c) \textbf{Simultaneous Construction of Clusters and
$\beta$-Bounded Exemplars} (\sccre)

\smallskip
\noindent
\underline{\textsf{Given:}}~ A set 
$X = \{x_1, x_2, \ldots, x_n\}$ of $n$ instances to be clustered,
integer $k$, where $2 \leq k \leq n$ (the number of clusters), 
a value $\epsilon > 0$ and
integer $\beta$ (upper bound the total number of exemplars for all clusters).

\smallskip
\noindent
\underline{\textsf{Requirement:}}~ 
Find a partition $X$ into at most $k$ clusters $C_1, C_2, \ldots, C_k$ 
and the corresponding
exemplar sets $\cale_1$, $\cale_2$, $\ldots$, $\cale_k$ 
as in the \scce{} problem above
with the requirements for compactness of clusters (Condition (i)), 
distinctness of explanation (Condition (iii)) but now:
\begin{itemize}
\item \underline{Upper Bound on the Number of Exemplars:}
(ii)~ $\sum_{j=1}^{k}|\cale_j| \leq \beta$, and

\item \underline{Relaxing the Condition that Every Instance be Explained:}
(v)~ The number of instances which have an exemplar in the same cluster
is as large as possible. 
\end{itemize}


\noindent
We will discuss an approximation algorithm for \sccre{} in 
Section~\ref{sec:main_prob_analysis}.
Note that compared to \scce{},
not every instance will be explained (i.e., covered by an exemplar). 
Instances which are not explained can be identified; 
this could be useful in that such instances may 
represent anomalies.

\section{Algorithmic Results}
\label{sec:main_prob_analysis}


\subsection{Finding a Minimum Set of Exemplars For A Single Cluster}
\label{sse:mse_hardness}

We begin with a complexity result for the Minimum Set of
Exemplars (\msec) problem for a single cluster.
As can be seen from its proof, 
this complexity result holds even when the
given set of instances $X$ consists of points in 
2D-Euclidean space.

\begin{theorem}\label{pro:mse_hardness}
The MSE problem is \cnp-hard even when the set of instances $X$
consists of points in 2D-Euclidean space and the distance
between any two points is their Euclidean distance.
\end{theorem}

\noindent
\textbf{Proof:}~ The proof is by a reduction
from the minimum dominating set (MDS) problem for unit disk
graphs discussed in Section~\ref{sse:addl_def}.
Let the MDS problem be specified by a unit disk graph $G(V,E)$,
where the radius of each disk is $r$,
and let $\beta \leq |V|$ be the given upper bound on the size of
a dominating set.
We construct a set of  instances $X$ for the MSE problem 
as follows.
For the disk corresponding to each vertex $v_i$, we create an instance
$x_i \in X$, where the coordinates of $x_i$ are those of the center of
the disk corresponding to $v_i$.
The exemplar distance $\epsilon$ is set to $2r$ and the bound on
the number of exemplars is set to $\beta$.
Clearly, this construction can be done in polynomial time.

Suppose $V'$ is a dominating set for $G$ with at most $\beta$ nodes.
We can show that the instances corresponding to the nodes in $V'$ form
the exemplar set \cale{} for $X$ as follows.
Consider any instance $x_j$ in $X$ which is not an exemplar.
Since $V'$ is a dominating set and the node $v_j$ corresponding to $x_j$
is not in $V'$, there is a node $v_i \in V'$
such that the edge $\{v_i,v_j\}$ is in $E$.
Since $G$ is a unit disk graph, the distance between the centers of
the disks corresponding to $v_i$ and $v_j$ is at most $2r$ which is
equal to $\epsilon$ by our construction; that is,
the distance between $x_j$ and the exemplar $x_i$ is
at most $\epsilon$.
Therefore, \cale{} is a set of exemplars of size at most $\beta$ for $X$.

Now, suppose \cale{} is a set of exemplars of size at most $\beta$ for $X$.
Let $V'$ be the set of nodes of $G$ corresponding to the instances in \cale.
We claim that $V'$ is a dominating set for $G$.
To see this, consider any node $v_j$ which is not in $V'$.
The instance $x_j$ corresponding to $v_j$ has an exemplar $x_i \in \cale$
and the distance between $x_i$ and $x_j$ is at most $2r$.
Since $G$ is a unit disk graph, the edge $\{v_i, v_j\}$ is in $E$.
In other words, $V'$ is a dominating set for $G$,
and this completes the proof. \QED

\subsection{An Approximation Algorithm for \scce}
\label{sse:scce_approx}

The \scce{} problem requires us to find a clustering where the diameter
of each cluster and the number of exemplars are as small as possible.
Since each of these problems is computationally intractable, we present
an algorithm that provides a provable performance guarantee for each
of these measures.

\medskip

\noindent
\textbf{Overview of the algorithm.}~
First, the algorithm takes the set $X$ and produces pairwise disjoint blocks $B_1$, $B_2$, $\ldots$, $B_k$
to minimize the maximum diameter \cite{Gon-1985} .
It then uses a greedy approximation algorithm for 
the Minimum Set Cover (MSC) problem \cite{Vaz-2001} to find a 
near-minimal set of exemplars $A$ for the set $X$. 
For each cluster $C_j$, the exemplar set $\cale_j$ is given by
$\cale_j = B_j \cap A$, $1 \leq j \leq k$.
Finally, each cluster $C_j$ consists of the exemplar set $\cale_j$ and
all the non-exemplars covered by $\cale_j$. 
This ensures that the exemplars are pairwise disjoint and
that each non-exemplar is covered by an exemplar in the same cluster.
Note that we only move \underline{non-exemplars} from their original blocks 
(i.e., $B_1$, $B_2$, $\ldots$, $B_k$) to new clusters (i.e., 
$C_1$, $C_2$, $\ldots$, $C_k$).
This is crucial to ensure the performance guarantee on
the maximum diameter. 
An outline of our approximation procedure is shown as
Algorithm~\ref{alg:scce_approx}. 
Note that if an instance $x$ is covered by multiple exemplars,
it can be assigned to \underline{any} cluster that has an exemplar for $x$.
The following theorem shows the performance guarantee
provides by  Algorithm~\ref{alg:scce_approx}.

\begin{algorithm}[tb]
\DontPrintSemicolon
\BlankLine
\SetKwInOut{Input}{Input}\SetKwInOut{Output}{Output}
\Input{
A set of instances $X$, the number of clusters $k$ and the
exemplar distance bound $\epsilon$.
}
\Output{
A clustering of $X$ into $k$ clusters and a set of
exemplars for each cluster to satisfy the requirements 
of the \scce{} problem.
}
\BlankLine
\textbf{Block Creation.}~ Use Gonzalez's approximation algorithm \cite{Gon-1985} 
to obtain $k$ (disjoint) blocks $B_1, B_2, \ldots, B_k$ of $X$. \;
\BlankLine
\textbf{Exemplar Neighborhood Set Construction.}~
For each $x_i \in X$, find $S_i$, the set of all instances $x_j \in X$
such that $d(x_i, x_j) \leq \epsilon$. 
(Thus, $x_i$ can serve as the exemplar for each instance in $S_i$.)\;
\BlankLine
\textbf{Exemplar Selection.}~ Construct the Minimum Set Cover (MSC)
problem consisting of the base set $X$ and the set collection
\bbs{} = $\{S_1, S_2, \ldots, S_n\}$.
Use a greedy approximation algorithm  for MSC \cite{Vaz-2001} to
construct a near-optimal set cover given by the 
subcollection $\bbs_1 \subseteq \bbs$.
Obtain the exemplar set $A$ as follows: for each $S_i \in \bbs_1$,
add $x_i$ to $A$.\;
\BlankLine
\textbf{Cluster Creation.}~ Create $k$ empty clusters $C_1, C_2, \ldots, C_k$.
\BlankLine
\textbf{Exemplar Assignment.}~ For each cluster $C_j$, the set $\cale_j$ of
exemplars is given by $\cale_j = B_j \cap A$.
Add $\cale_j$ to $C_j$.\;
\BlankLine
\textbf{Non-Exemplar Assignment.}~ Consider each cluster $C_j$.
For each exemplar $x_i \in C_j$, add each instance in $S_i - A$ 
(i.e., each non-exemplar in $S_i$) to $C_j$.\;
\BlankLine
Output the set of clusters $C_1$, $C_2$, $\ldots$, $C_k$
and the corresponding exemplars\newline $\cale_1$, $\cale_2$, $\ldots$, $\cale_k$.
\caption{Approximation Algorithm for \scce}
\label{alg:scce_approx}
\end{algorithm}

\begin{theorem}\label{thm:perf_ssce_approx}
The solution produced by 
Algorithm~\ref{alg:scce_approx} satisfies the following properties:
(i) The diameter of each cluster is at most $2(D^* + \epsilon)$, where
$D^*$ is the optimal diameter for a $k$-clustering of $X$ and $\epsilon$
is the exemplar distance.
(ii) Every instance in $X$ has an exemplar 
within the same cluster.
(iii) The sets of exemplars for the $k$ clusters are pairwise disjoint.
(iv) The total number of exemplars generated 
by the algorithm is at most $O(N^*\log{n})$, where $N^*$ is the minimum
number of exemplars needed to cover all the instances in $X$.
\end{theorem}

\noindent
\textbf{Proof:}~ To prove Part~(i), we first note that the approximation
algorithm used in Step~1 guarantees that the maximum diameter of the
clusters produced in that step is at most $2D^*$, where $D^*$ is the
optimal solution value for $X$.
Step~6 of the algorithm moves only non-exemplars between clusters.
We need to show that after these moves, the maximum diameter is
at most $2(D^*+\epsilon)$.
To see this, consider any cluster $C_i$ and any pair of instances 
$x_a$ and $x_b$ in $C_i$.
There are three cases to consider.

\smallskip
\noindent
\underline{Case 1:}~ Both $x_a$ and $x_b$ are exemplars.
In this case, both $x_a$ and $x_b$ must be in $B_i$
since we chose $\cale_i = B_i \cap A$.
Thus, at the end of Step~1, $d(x_a, x_b) \leq 2D^*$.

\smallskip
\noindent
\underline{Case 2:}~ 
One of them, say $x_a$,  is an exemplar and the other (i.e., $x_b$)
is a non-exemplar that got moved into $C_i$. 
In this case, $C_i$ contains an exemplar $x_q$ at a distance of 
at most $\epsilon$ from $x_b$.
Since $d(x_a, x_q) \leq 2D^*$ and $d(x_q, x_b) \leq \epsilon$, it
follows from triangle inequality that $d(x_a, x_b) \leq 2D^*+\epsilon$.

\smallskip
\noindent
\underline{Case 3:}~ 
Both $x_a$ and $x_b$ are non-exemplars which were moved into $C_i$.
In this case, $C_i$ contains exemplars $x_p$ and  $x_q$ such that
$d(x_a, x_p) \leq \epsilon$ and $d(x_b, x_q) \leq \epsilon$. 
Further, $d(x_p, x_q) \leq 2D^*$.
Now, using triangle inequality, it follows that 
$d(x_a, x_b) \leq 2(D^* + \epsilon)$, and this completes
our proof of Part~(i).

The result in Part~(ii) follows since the set $A$ 
constructed in Step~3 is an exemplar set for $X$ and 
each non-exemplar instance $x_j$ gets moved (in Step~6) to a cluster   
containing an exemplar for $x_j$.
Since the blocks constructed in Step~1 are pairwise disjoint, so are the exemplar
sets constructed in Step~5; this proves Part~(iii).
Since Step~3 uses the greedy approximation algorithm for MSC and
this algorithm provides a performance guarantee of $O(\log{n})$ \cite{Vaz-2001}, 
the total number of exemplars produced in Step~3 
is at most $O(N^* \log{n})$,
where $N^*$ is the minimum number of exemplars needed to cover
all instances in $X$.
This establishes Part~(iv) and the theorem follows. \QED

\medskip

\noindent
\textbf{Remark:} 
Since Step~3 in Algorithm~\ref{alg:scce_approx}
uses an approximation algorithm for MSC, the performance guarantee with respect to
the number of exemplars is $O(\log{n})$, where $n = |X|$.
Theoretically, one can get a better approximation by transforming
the Exemplar Selection steps (i.e., Steps 2 and 3 of the algorithm)
into that of finding a near-optimal 
dominating set for unit disk graphs 
in an Euclidean space whose dimension $\ell$ is the same 
as that of the points in $X$.
This is done by placing an $\ell$-dimensional ball 
of radius $\epsilon/2$ at each instance in $X$.
The corresponding unit disk graph has a node for each instance in $X$ 
and there is an edge between two nodes if the corresponding 
balls intersect or touch.
It can be verified that any dominating set for this graph 
provides the necessary set of exemplars.
An approximation scheme which provides a performance guarantee of $(1+\delta)$
for any fixed $\delta > 0$ is known for the minimum dominating set problem
for such graphs \cite{HM-1985}.
Thus, one can obtain a performance guarantee of
$(1+\delta)$ for any fixed $\delta > 0$ with respect to the number of exemplars.
However, this approximation scheme is impractical even for data sets of 
moderate size since its running time has the factor $O(n^{(1/\delta)^2})$. 
(Thus, even when $\delta = 0.5$, the running time has the factor $O(n^4)$.)
For this reason, we decided to use the 
MSC-based approximation algorithm in our experiments.

\medskip

\noindent
\textbf{Running time of Algorithm~\ref{alg:scce_approx}:}~
We can estimate the asymptotic running time this approximation
algorithm as follows.
Step~1 uses Gonzalez's algorithm which has a running time of $O(nk)$,
where $n$ is the number of instances and $k$ is the number of clusters
\cite{Gon-1985}.
Step~2 constructs the neighborhood set for each instance and
can be done in time $O(n^2)$.
Step~3 runs the greedy set cover heuristic for which the running
time is $O(W)$, where $W$ is the sum of the sizes of all the
sets \cite{Blelloch-etal-2012}.
In our case, since there are $n$ sets and each set is of size at most $n$,
$W \leq n^2$; that is, Step~3 runs in time $O(n^2)$.
Step~4 runs in $O(k)$ time.
Using a bit vector representation for each set, Steps~5 and 6 can be
implemented to run in time $O(nk)$.
Since $k \leq n$, the overall running time of 
Algorithm~\ref{alg:scce_approx} is $O(n^2)$.

\subsection{An Approximation Algorithm for \sccre}
\label{sse:sccre_approx}


When $\epsilon$ is small, our approximation algorithm 
for  \scce{}  generates a solution with a small cluster 
diameter; however, it may yield a large number of exemplars
leading to an overly complicated explanation.  The goal
of \sccre{} is also to find a clustering with a small maximum
diameter but we relax the requirement to have exemplars for
\textbf{all} the instances.  Instead, we are given an upper bound
on the total number of exemplars for all clusters, and we want to maximize
the number of instances with exemplars subject to
the bound on the number of exemplars.

We now present an approximation algorithm that provides 
a provable performance guarantee for the diameter as well
as the number of instances covered by exemplars in each cluster.
This algorithm is similar to the one for the \scce{} problem
(Algorithm~\ref{alg:scce_approx}) except that it
uses a known approximation algorithm for the Budgeted Maximum
Coverage (BMC) problem \cite{Khuller-etal-1999} in Step~3
instead of the approximation algorithm for the MSC problem. 
The steps of this approximation algorithm are shown as
Algorithm~\ref{alg:sccre_approx}.
The following theorem establishes the performance guarantee provided
by the Algorithm~\ref{alg:sccre_approx}.

\begin{algorithm}[tb]
\DontPrintSemicolon
\SetKwInOut{Input}{Input}\SetKwInOut{Output}{Output}
\Input{
A set of instances $X$, the number of clusters $k$, the
exemplar distance bound $\epsilon$ and an upper bound $\beta$ on
the total number of exemplars for all clusters.
}
\Output{
A clustering of $X$ into $k$ clusters and a set of
exemplars for each cluster to satisfy the requirements 
of the \sccre{} problem.
}
\BlankLine
\textbf{Block Creation.}~ Use Gonzalez's approximation algorithm \cite{Gon-1985} 
to obtain $k$ (pairwise disjoint) blocks $B_1, B_2, \ldots, B_k$ of $X$. \;
\BlankLine
\textbf{Exemplar Neighborhood Set Construction.}~
For each $x_i \in X$, find $S_i$, the set of all instances $x_j \in X$
such that $d(x_i, x_j) \leq \epsilon$. 
(Thus, $x_i$ can serve as the exemplar for each instance in $S_i$.)\;
\BlankLine
\textbf{Exemplar Selection.}~ Construct the Budgeted Maximum Coverage (BMC)
problem consisting of the base set $X$, the set collection
\bbs{} = $\{S_1$, $S_2$, $\ldots$, $S_n\}$ and the budget $\beta$.
Use the greedy approximation algorithm  for BMC \cite{Khuller-etal-1999} to
construct a subcollection $\bbs_1 \subseteq \bbs$.
Obtain the exemplar set $A$ as follows: for each $S_i \in \bbs_1$, 
add $x_i$ to $A$.\;
\BlankLine
\textbf{Cluster Creation.}~ Create $k$ empty clusters $C_1, C_2, \ldots, C_k$.
\BlankLine
\textbf{Exemplar Assignment.}~ For each cluster $C_j$, the set $\cale_j$ of
exemplars is given by $\cale_j = B_j \cap A$.
Add $\cale_j$ to $C_j$.\;
\BlankLine
\textbf{Non-Exemplar Assignment.}~ Consider each cluster $C_j$.
For each exemplar $x_i \in C_j$, add each instance in $S_i - A$ 
(i.e., each non-exemplar in $S_i$) to $C_j$.
The set of instances $X'$  which don't have exemplars is given by 
$\displaystyle{X' = X - \cup_{S_i \in \mathbb{S}_1} S_i}$.\;
\BlankLine
Output the set of clusters $C_1$, $C_2$, $\ldots$, $C_k$
and the corresponding exemplars\newline 
$\cale_1$, $\cale_2$, $\ldots$, $\cale_k$.
\caption{Approximation Algorithm for \sccre}
\label{alg:sccre_approx}
\end{algorithm}

\begin{theorem}\label{thm:perf_sscre_approx}
The solution produced by
Algorithm~\ref{alg:sccre_approx} satisfies the following properties:
(i) The diameter of each cluster is at most $2(D^* + \epsilon)$, where
$D^*$ is the optimal diameter for a $k$-clustering of $X$ and $\epsilon$
is the exemplar distance.
(ii) The sets of exemplars for the $k$ clusters are pairwise disjoint.
(iii) The total number of instances with exemplars is
at least $(1-1/\mathrm{e})Q^*$, where $\mathrm{e}$
is the base of the natural logarithm and
$Q^*$ is the maximum number of instances in $X$ that can have 
exemplars under the constraint
that the total number of exemplars is at most $\beta$.
\end{theorem}

\noindent
\textbf{Proof:}~ The proofs of Parts (i) and (ii) of the
theorem are identical to the ones given in the proof of 
Theorem~\ref{thm:perf_ssce_approx}.
Part~(iii) follows from \cite{Khuller-etal-1999}
that the greedy approximation
algorithm for BMC covers at least $(1-1/\mathrm{e})Q^*$ elements,
where $Q^*$ is the maximum number of elements that can be covered 
using at most $\beta$ sets. \QED

\medskip

\noindent
\textbf{Running time of Algorithm~\ref{alg:sccre_approx}:} 
The estimation of the asymptotic running time of
Algorithm~\ref{alg:sccre_approx} is similar to that of
Algorithm~\ref{alg:scce_approx}.
The main difference between the two algorithms is that while
Algorithm~\ref{alg:sccre_approx} uses the greedy algorithm for
the BMC problem in Step~3 while 
Algorithm~\ref{alg:scce_approx} uses the greedy algorithm for
the Minimum Set Cover (MSC) problem.
However, the asymptotic running time of the greedy algorithm for BMC
is also the same as that of the greedy algorithm 
for MSC \cite{Khuller-etal-1999}. 
Therefore, the running time of 
Algorithm~\ref{alg:sccre_approx} is also $O(n^2)$.

\section{Experiments}\label{sec:experiments}


Code and data to reproduce and document the experiments are 
available\footnote{URL: \textcolor{blue}
{\url{www.cs.ucdavis.edu/~davidson/SCCE-DMKD-main.zip}}.
All code and public data are located at the site.} 
with the exception of the Harry Potter novel
data which is not in the public domain but is freely available.   We
have tried to quantitatively and qualitatively evaluate our approach's
usefulness for explanation to a human. We explore several directions
including generating summaries of a novel which we compare against
human written summaries. Similarly, we explore quantitative measures
on human faces in the wild data, and for completeness, a qualitative
analysis of a standard digit data set. Finally, in an emerging direction of
using explanation for machines (not humans), we explore using exemplars
for SVM transfer  learning. 

\medskip

\noindent
\textbf{Time Complexity.} Our approximation algorithms run in
polynomial time (more precisely, in $O(n^2)$ time in the worst-case)
and have strong performance guarantees in terms of clustering quality
and explanation complexity. The run times for our algorithms are
as expected not as fast as simple $k$-means style algorithms but
our work comes with performance guarantees with respect to optimal
solutions and are much
faster than state of the art domain specific methods. For
example in our work on explaining deep embeddings for text
(Section~\ref{sse:text_data}), our \scce{} and \sccre{} algorithms
took 93 and 96 seconds respectively whilst the state of the art
method took 700+ seconds and $k$-means style algorithms (which lack
explanation) took under 10 seconds. Our algorithm has just two parameters,
namely $k$ and $\epsilon$, where the latter parameter naturally trades off
clustering quality and explanation complexity.

\subsection{Qualitative Experiments on Digits Data}
Here, we take the standard MNIST data set consisting of 10,000 written
digits. We embed them using tSNE \cite{maaten2008visualizing} and
use our algorithm to cluster them and generate exemplars. Our hope
is that the exemplars will be a varied representation of the different
ways of writing each digit. Through experiments, we empirically
verify how useful exemplars from text and images are from a predictive
perspective, but here visually inspect them for usefulness.  The
clusters found by our methods and approximate centroids (not exemplars)
are shown in Figure~\ref{fig:mnist}. 
(A larger version of the figure is 
given in Section~\ref{app:sse:larger_fig} of the Appendix.) 
For each cluster, we present the exemplars found in 
Table~\ref{tab:mnistExemplars}. Of course, the clustering does not have
100\% accuracy but we see that for well separated clusters (0, 5, 6, 7, 8 and 9),
the exemplars do indeed capture a variety of ways that the
digits are written. Quite surprisingly, many are fundamentally different
from the centroid. Take for example the digit 7. The centroid has the top
line pointing downwards but the exemplars show examples where the
top line is up and the vertical line is crossed. The exemplars vary
 by their form and also in the pressure applied to the pen.

\begin{figure*}
\includegraphics[scale=0.215]{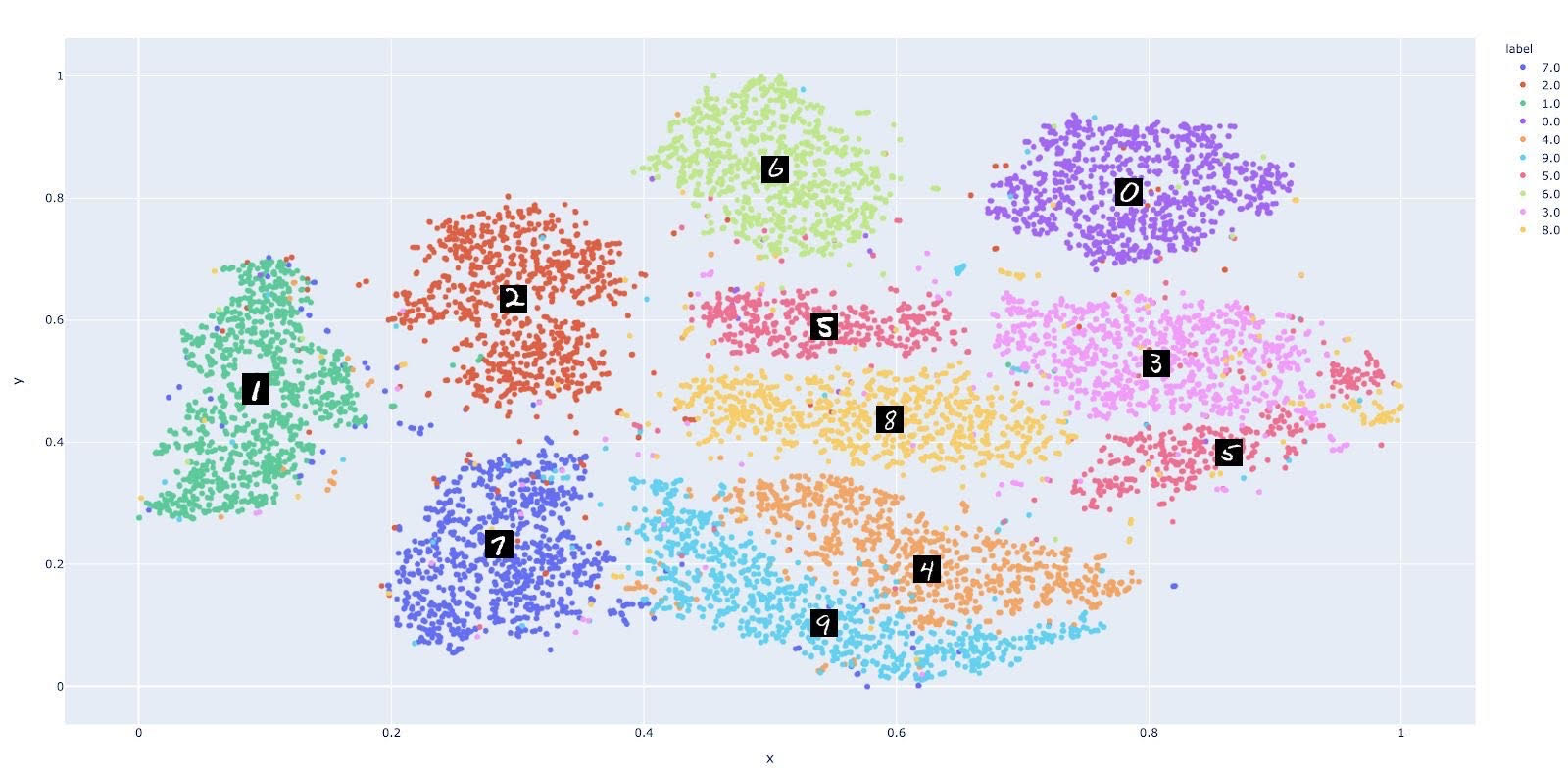}
\vspace{-0.15in}
\caption{Clusters and centroids (not exemplars) found by our 
method when applied to the MNIST dataset.  A larger version of the figure is 
given in Section~\ref{app:sse:larger_fig} of the Appendix.} 
\label{fig:mnist}
\vspace*{-0.1in}
\end{figure*}

\begin{table}
\begin{center}
\begin{tabular}{|c|c|c|c|}  \hline
0 & \includegraphics[scale=0.5]{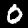} \includegraphics[scale=0.5]{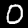} \includegraphics[scale=0.5]{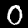} \includegraphics[scale=0.5]{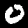} \includegraphics[scale=0.5]{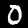} \\ \hline
1 & \includegraphics[scale=0.5]{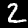} \includegraphics[scale=0.5]{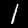} \includegraphics[scale=0.5]{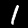} \includegraphics[scale=0.5]{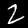} \includegraphics[scale=0.5]{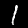}  \includegraphics[scale=0.5]{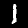} \includegraphics[scale=0.5]{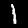}  
 \includegraphics[scale=0.5]{MNIST_EXEMPLARS/6_2.png} \includegraphics[scale=0.5]{MNIST_EXEMPLARS/6_3.png} \includegraphics[scale=0.5]{MNIST_EXEMPLARS/6_4.png} \\ \hline
2 & \includegraphics[scale=0.5]{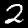} \includegraphics[scale=0.5]{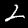} \includegraphics[scale=0.5]{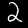} \includegraphics[scale=0.5]{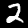} \includegraphics[scale=0.5]{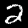}  \includegraphics[scale=0.5]{MNIST_EXEMPLARS/6_0.png}  \includegraphics[scale=0.5]{MNIST_EXEMPLARS/6_1.png} \\ \hline
3 & \includegraphics[scale=0.5]{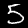} \includegraphics[scale=0.5]{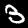} \includegraphics[scale=0.5]{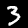} \includegraphics[scale=0.5]{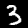} \includegraphics[scale=0.5]{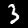} \\ \hline
4 & \includegraphics[scale=0.5]{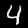} \includegraphics[scale=0.5]{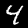} \includegraphics[scale=0.5]{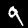} \includegraphics[scale=0.5]{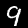} \includegraphics[scale=0.5]{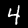} \\ \hline
5 & \includegraphics[scale=0.5]{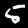} \includegraphics[scale=0.5]{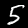} \includegraphics[scale=0.5]{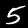} \includegraphics[scale=0.5]{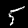} \includegraphics[scale=0.5]{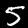} \\ \hline
6 & \includegraphics[scale=0.5]{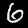} \includegraphics[scale=0.5]{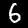} \includegraphics[scale=0.5]{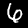} \includegraphics[scale=0.5]{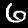} \includegraphics[scale=0.5]{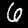} 
 \includegraphics[scale=0.5]{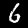} \includegraphics[scale=0.5]{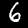}  
  \includegraphics[scale=0.5]{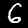} \includegraphics[scale=0.5]{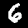} \includegraphics[scale=0.5]{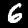} \\ \hline
7 & \includegraphics[scale=0.5]{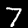} \includegraphics[scale=0.5]{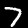} \includegraphics[scale=0.5]{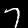} \includegraphics[scale=0.5]{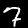} \includegraphics[scale=0.5]{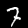} \\ \hline
8 & \includegraphics[scale=0.5]{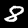} \includegraphics[scale=0.5]{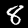} \includegraphics[scale=0.5]{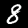} \includegraphics[scale=0.5]{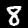} \includegraphics[scale=0.5]{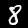} \\ \hline
9 & \includegraphics[scale=0.5]{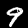} \includegraphics[scale=0.5]{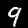} \includegraphics[scale=0.5]{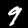} \includegraphics[scale=0.5]{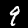} \includegraphics[scale=0.5]{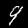} \\ \hline
\end{tabular}
\end{center}
\caption{The clusters and exemplars found by our method on the MNIST
data set. Note that the exemplars provide a variety of ways that
the digits are written and most importantly are quite different
from the centroids shown in Figure \ref{fig:mnist}.}
\label{tab:mnistExemplars}
\vspace*{-0.15in}
\end{table}

\subsection{Quantitative Experiments on Textual Data}\label{sse:text_data}

In this section, we evaluate the ability of exemplars to simplify a
corpus by  summarizing  content. In particular, we take the sentences
in the first Harry Potter (HP) book, embed them using deep learning,
apply Algorithms \ref{alg:scce_approx} and \ref{alg:sccre_approx} and concatenate the resultant exemplars
to form a summary. This is compared with a ranking based 
approach \cite{erkan2004lexrank,mihalcea2004textrank}
which can be viewed as choosing exemplars from a list based on
importance. These ranking methods are known to produce superior
results for HP books \cite{hattasch2020summarization} compared to
recent methods. We measure results by comparing against four
human written summaries.\footnote{\url{www.britannica.com},
\url{en.wikipedia.org}, \url{harrypotter.fandom.com},
\url{content.time.com}} Results (Table \ref{tab:rouge}) show that our
method performs better than these ranking method by 12.8\% and the
baseline of random selection of sentences by over 20\%. Most
importantly, our method's summary score is almost comparable (on
average) to the similarity between the human summaries themselves
(Table \ref{tab:rouge}).

We measure performance using the ROUGE score \cite{lin2002manual}
which is a standard method of evaluating the similarities between
computer generated summaries and human written summaries.  We
represent each sentence in the first HP book using the state-of-the-art
language model BERT \cite{devlin2019bert}.  Hence, the exemplars
generated by our method will be sentences in the book.  Specifically,
we fine-tune a pre-trained BERT-base model
(\url{https://huggingface.co/}) in two steps. First, we
add to the vocabulary terms words that are unique to the Harry
Potter universe (e.g., ``quidditch’’) and train the model with a
very low learning rate. Then, we fine-tune the model to produce a
relevant sentence embedding using the Sentence-BERT architecture
\cite{reimers2019sentence} to create a HP Specific BERT model.
\emph{It is important to note that all methods and baselines
use this embedding scheme.}

We compare our two methods against two approaches.
The first one is a random subset of sentences used as a control.
We repeat this random selection process 20 times.  The second
baseline is the widely used ranking approach for extracting summaries
\cite{mihalcea2004textrank,erkan2004lexrank}. These methods require
a graph which we construct from pairwise cosine similarities using
the sentence embedding obtained with our fine-tuned BERT model.
This is a time tested method with thousands of
citations and in 2020 still produces state of the art results for
the HP literature \cite{hattasch2020summarization}. 
For all methods except for \scce,
we fix the number of sentences extracted to be equal to the number
of sentences in the ground truth summaries.  We use
6 clusters chosen after hyper-parameter tuning to find the stablest
clusters. 
An example summary is shown in Section~\ref{app:sse:hp_explanation}
of the  Appendix. 


\begin{table*}
\begin{tabular}{|llllll||ll|}
\hline
          & \multicolumn{5}{c||}{Methods} & \multicolumn{2}{c|}{Relative Performance} \\ 
          & \scce{}   & \sccre{}  & Ranking & Random & Other   & To  & To Other  \\
          & (Ours) & (Ours) &\cite{mihalcea2004textrank}& & Summaries   
                                    & Ranking & Summaries\\ \hline
Sum-1 & 31.65       & \textbf{33.81}                & 28.86    & 23.69  & 38.09                   & +17\%     & -11\%   \\
Sum-2 & 29.58       & \textbf{31.26}                & 28.73    & 25.89       & 25.82 
& +9\%     & +21.1\%   \\
Sum-3 & 27.08       & \textbf{28.78}                & 26.58    & 22.68    & 31.32 & +8.3\%     & -8.1\%   \\
Sum-4 & 33.33       & \textbf{34.11}                & 28.31    & 24.29        & 36.08                 & +17\%     & -5.5\%  \\ \hline
\end{tabular}
\caption{The ROUGE-1 F1-scores (the larger the better) measuring
the similarity of our two methods, one state of the art (Ranking),
one baseline (Random) to four human written summaries (one per row)
of the first Harry Potter novel. For each summary, we also
report the the average similarity to the remaining three summaries
(Human Summaries). Each computational method (except \scce{})
generates the same number of sentences as the summary against
which it is compared.}

\label{tab:rouge}
\end{table*}

\begin{table}
\begin{center}
\begin{tabular}{|c|c|}\hline
\textbf{Clustering Artifact Used} & \textbf{Accuracy} \\ \hline\hline
 Exemplars & 48.33 \\  \hline
 Cluster centers & 44.00 \\\hline
 All Points & 42.00 \\\hline
 Random Points in cluster & 44.66 \\   \hline
\end{tabular}
\end{center}
\caption{Measuring the effectiveness of exemplars to explain/predict
a person from images. Competing methods use the  same clustering
we find but instead use $k$-Nearest-Neighbor  for prediction with
different aspects/artifacts of the cluster.
The value $\epsilon$ is tuned and set to 0.6 to 
maximize the stability of clusters.}
\label{simulatedImages}
\end{table}

\subsection{Quantitative Experiments on Facial Data}
One way to determine whether an explanation is useful is to check if it helps a
human to understand the underlying concepts which are the clusters. A
typical test of exemplar theory given to humans \cite{murphy2004big} 
is the task of identifying several people they have never seen
before using only a small set of exemplars of the people. We make the task
challenging by choosing three similar men (Gerhard Schr\"{o}der,
Jacques Chirac and Tony Blair) and use just 40
images of each person, with half used for clustering and half for testing.

\begin{figure*}
\includegraphics[scale=0.65]{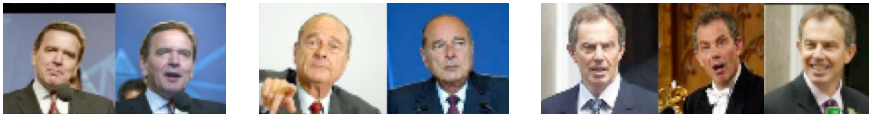}
\vspace{-4.7in}
\caption{Faces in the Wild Experiments. Exemplars found for our three clusters correspond to the three people used in this experiment, Gerhard Schröder (left), Jacques Chirac (middle) and Tony Blair (right). Note the exemplars of the same person differ mainly by the position of the mouth.} 
\label{fig:faceExemplars}
\end{figure*}
For reproducibility, we simulate a person by the most simple learning algorithm, namely $k$-nearest
neighbor ($k$-NN). We cluster images of three well-represented individuals
from the Labeled Faces in the Wild dataset \cite{huang2008labeled} using our method.
Images are first processed into embeddings via FaceNet, a deep
embedding network.
After clustering using our method, each cluster was assigned the label of its most 
well-represented individual.

We created three baselines to predict the person in the hold out image: 1)
Using a nearest centroid approach,
2) Using a $k$-NN approach with all points and 3) Using a $k$-NN approach but with random 20\% of points from each cluster.
After conducting this experiment five times with five different
training/testing splits, we obtained the results summarized in 
Table~\ref{simulatedImages}. This experiment demonstrates that exemplars
produced by our method are more useful than \textbf{other artifacts} of the very same clustering namely centroids, all points and random subsets of points.
A possible reason for the improvement is that our method chooses
a more diverse collection of instances (Figure~\ref{fig:faceExemplars}).

\subsection{Exemplars for Instance Transfer Learning}
Our exemplar and clustering discovery method can also help to explain a problem to a
machine.  Essentially, our method identifies clusters of points and
important examples of each cluster. Here we use those important
points to do instance transfer learning for support vector machines
(SVMs). Transfer learning uses a source task to help a target
task.  We use the well known pendigits dataset \cite{alimouglu2001combining}
to transfer the task of predicting between two digits to help another
task of predicting between two very similar digits.  For example,  we can learn the source task of 
\texttt{1 vs 9} and transfer it to help the \texttt{1 vs 7} task as shown in Table \ref{tab:SVM}.

Recall that with a SVM the vector \textbf{w} implicitly defines the hyperplane and a constraint to separate the two classes is defined as shown below in equation \ref{eq:SVM}.  A common method of performing SVM transfer learning is to add an additional
constraint to the problem that requires the hyperplane to also separate the classes in the source problem.
Note the last constraint in equation \ref{eq:SVM} contains the transfer as $(x^s_i,y^s_i)$
are the support vectors from the previously solved SVM for the
source problem.  In our experiments,  rather than transferring over these support vectors,
we can instead transfer the exemplars.  We use the bounded version
of our formulation to transfer over the same number of instances
as support vectors in the source problem.  Results 
in Table~\ref{tab:SVM}  show promise and a future 
direction of exemplars augmenting existing ML tasks.
\begin{equation}
\mathrm{argmin}_{\textbf{w},\,w_0}  \frac{1}{2}||\textbf{w}||^2  \\
\label{eq:SVM}
\end{equation}
\begin{center}
\begin{tabular}{llcll}
s.t. & $y_i(\mathbf{w}^T.\mathbf{x}_i + w_0)$     &$\ge$ & $+1~~\forall i$~~~ and \\
     & $y^s_j(\mathbf{w}^T.\mathbf{x^s_j} + w_0)$ &$\ge$ & $+1~~\forall j$ 
\end{tabular}
\end{center}

\begin{table} [t]
\begin{center}
\begin{tabular}{|p{1.0in}|p{0.8in}|p{0.7in}|p{0.7in}|}
\hline
\textbf{(Source) Target} & \textbf{No Transfer} & \textbf{Transfer Support Vectors} & \textbf{Transfer Exemplars} \\ \hline
(1 vs 9) 1 vs 7 & 0.79 & 0.83 & 0.91 \\
(2 vs 8) 2 vs 3 & 0.78 & 0.81 & 0.89 \\
(3 vs 8) 3 vs 9 & 0.81 & 0.84 & 0.88 \\
(1 vs 7) 1 vs 9 & 0.82 & 0.83 & 0.90 \\
(2 vs 3) 2 vs 8 & 0.80 & 0.81 & 0.91 \\
(3 vs 9) 3 vs 8 & 0.73 & 0.70 & 0.89 \\
 \hline \hline
 (1 vs 9) 1 vs 7 & 0.63 & 0.72 & 0.80 \\
(2 vs 8) 2 vs 3 & 0.64 & 0.73 & 0.81 \\
(3 vs 8) 3 vs 9 & 0.66 & 0.72 & 0.81 \\
(1 vs 7) 1 vs 9 & 0.62 & 0.71 & 0.79 \\
(2 vs 3) 2 vs 8 & 0.61 & 0.69 & 0.83 \\
(3 vs 9) 3 vs 8 & 0.59 & 0.63 & 0.77 \\ 
\hline
 \end{tabular}
 \end{center}
\caption{Accuracy for Transfer Learning. 350
training instances of each digit were randomly chosen for both
source and target problems. The 3rd column shows transferring the
support vectors and the 4th column shows transferring the exemplars
from our work. Results are averaged over 100 random trials. Results
above (below) the double lines use all 8 pairs (first 4 pairs) of
coordinates. Using just half the features produces nearly twice
as many support vectors.} 
\label{tab:SVM} 
\end{table}

\section{Related Work}
\label{sec:related}

\textbf{Explanation and Clustering.} The machine learning community 
has studied explaining clusters from
two perspectives. The one-view approach of conceptual clustering
\cite{michalski1983learning,frost2020exkmc,saisubramanian2020balancing} 
proposes a task that is similar to our own 
(i.e., finding a clustering and its description), but requires
that the features used to perform clustering are human interpretable. 
This work can be seen as expounding the definition-based
theory of concepts \cite{murphy2004big} as it defines the 
concept explicitly using an underlying language.
Two-view work attempts to find a clustering using one set of 
features and a description using another set of features 
(typically human interpretable tags) 
such as our own \cite{dao2018descriptive}. This work again
takes a descriptive (non-exemplar) based approach to explanation
and most importantly does not scale beyond a few hundred points 
as it performs  Pareto optimization.
In contrast, the work presented in this paper scales to hundreds of
thousands of points; all our experiments run on standard laptops in under a minute.
More recent work \cite{davidson2018cluster,sambaturu2020efficient} 
has explored explaining
a \emph{given} clustering using a set of auxiliary tags;
it does not find a clustering itself. 

\smallskip
\noindent
\textbf{Concept Theory.} A motivation for our work comes from human psychology and philosophy, and
in particular, the large body of work known as theories of concepts
\cite{murphy2004big}. This field defines the building
blocks of knowledge as concepts but there are 
several different definitions of concepts. 
Our work falls under the exemplar
based theory  which defines a concept by exemplars of that concept
whilst the competing classical definition theory requires defining
properties of the concept 
(e.g., ``Jaques Chirac is tall, bald and cherubic faced"). 
However, this classic definition
requires the instances to be described using an understandable set of
features which is not the case for deep embeddings or most
complex domains. Though exemplar based theories suggest how to
define a concept, the problem of finding/choosing those exemplars
is not well posited by the psychology community. Our work can
be seen as formalizing and extending this theory by formulating
simultaneous concept and exemplar discovery as a combinatorial
optimization problem.

\medskip

\noindent
\textbf{Comparison to DBSCAN and Other Density Based Clustering
Methods.} Superficially, our method may seem to be similar to  DBSCAN
\cite{Ester-etal-1996} and other similar algorithms as it uses notions such as
$\epsilon$-neighbors. However, there are several fundamental
differences. Firstly, our method is guaranteed to use the specified number
or near-minimum number of exemplars, where as DBSCAN, while being a
very useful method, does \underline{not provide such guarantees}.
Similarly, our method has an explicit clustering objective (i.e., to minimize
the maximum cluster diameter) where as  DBSCAN does not. Finally,  DBSCAN
is a not designed so that the core points can be considered explanations
of the clusters. As a consequence, it is not meaningful to 
compare our method with DBSCAN.

\medskip

\noindent
\textbf{Comparison to Multiple Centroid Methods.} 
An area that is superficially similar to our own work is 
finding multiple centroids per cluster; these centroids are
sometimes referred to exemplars. However,
there are significant differences with respect to the 
definition of an exemplar,  the purpose of the exemplars 
and the efficiency of the algorithms. 

The multi-centroid/exemplar methods are
specifically focused on identifying multiple centroids in each
cluster, where each centroid specifies a new sub-cluster  
(e.g., \cite{MEAP-Wang-etal-2013,KMEAP-Wang-Chen-2014}).
While these methods allow a
user to specify the number of clusters $k$,
the algorithms may find more clusters, that is,  possible sub-clusters
within each cluster \cite{KMEAP-Wang-Chen-2014}.
One can view these are finding a one layer hierarchy 
within each cluster and experimental results typically
compare these algorithms  against  hierarchical clustering methods.

In our work, an exemplar has a very precise definition: namely a
point $x$ is an exemplar for another point $y$ if and only if $x$ is within a
certain distance from $y$.  The work on multiple centroid clustering
has no such definition.
Further, the exemplars generated by our methods 
are motivated by the need to explain clusters rather than
to identify sub-clusters and hence yield fundamentally different results.
As an illustrative example, consider a cluster with 
points uniformly distributed throughout it.  
Methods such as MEAP and K-MEAP
\cite{MEAP-Wang-etal-2013,KMEAP-Wang-Chen-2014} will return just one
exemplar for the entire cluster, as there are no distinct sub-clusters.
However, our methods will return multiple exemplars 
when $\epsilon$ is small enough.  
Figure \ref{fig:mnist} provides such an example where the 
clusters are tightly defined with no sub-clusters.
Finally, while the methods in
\cite{MEAP-Wang-etal-2013,KMEAP-Wang-Chen-2014} 
provide no formal performance
guarantees with respect to either of the two objectives
considered in our work (i.e.,
the cluster quality and the number of exemplars chosen), 
our methods have provable performance guarantees for both
of the objectives.

\section{Future Work and Conclusions}\label{sec:concl}

XAI for clustering is an under-studied problem compared
to supervised learning. Here we explore a style of explainable-by-design
algorithm that simultaneously finds clusters and exemplars to
describe those clusters. The idea of using exemplars has several
benefits. Firstly, it has pedagogic benefits in that humans are
known to naturally understand concepts in terms of exemplars
\cite{murphy2004big}. How humans naturally cluster and then organize
these exemplars into hierarchical structures will motivate future
work. Secondly, the use of exemplars is perhaps the only way to
explain data when it is clustered in high dimensional uninterpretable spaces
such as deep embeddings.  We show that finding a small set of exemplars for just
one cluster is \cnp-hard and design approximation
algorithms with provable performance guarantees. 
We demonstrate their usefulness in four tasks: (i) to
generate a summary of a book which is compared to a human summary,
(ii) to generate exemplars for the classic MNIST data set, (iii)
to generate exemplars that can be used to identify people and (iv)
to perform instance transfer learning.  Our approach is based on
classic computations (e.g., minimum set cover) 
but the combination of the methods is novel.
This has the advantage of being able to leverage known results and
implementations of these classic algorithms; see code in the 
following repository:
\newline\textcolor{blue}
{\url{www.cs.ucdavis.edu/~davidson/SCCE-DMKD-main.zip}}.
This has other advantages such as ease of
parallel implementation.  Like most ML methods, our methods also need parameter
tuning. Most clustering algorithms need to tune $k$ (the number of clusters) and
our method adds another parameter $\epsilon$ (the coverage of an
exemplar). The relationship between $\epsilon$ and the number of
exemplars allows for a natural trade off between the complexity of
the explanation and cluster compactness as per our bounds.  
If the data to be clustered is human interpretable, then other methods
of explanation are also suitable \cite{frost2020exkmc,michalski1983learning} but
exemplars are a natural and pragmatic way to explain complex data.

\medskip

\noindent
\textbf{Acknowledgments:}~
This work was supported in part by NSF Grants IIS-1908530 and IIS-1910306 titled:
``Explaining Unsupervised Learning: Combinatorial Optimization Formulations,
Methods and Applications''.


\bibliographystyle{spmpsci}      
\bibliography{refs}


\clearpage

\begin{center}
\dunder{\Large{\textbf{Appendix}}} 
\end{center}

\section{Additional Material for Section~\ref{sec:experiments}}

\subsection{Harry Potter Explanations By Our Method}
\label{app:sse:hp_explanation}

Here we present the explanation generated by our approach. We color
code the exemplars by the cluster they belong to.

\bigskip

\textcolor{orange}
{At that moment the telephone rang and Aunt Petunia
went to answer it while Harry and Uncle Vernon watched Dudley unwrap
the racing bike a video camera a remote control airplane sixteen
new computer games and a VCR.} 
\textcolor{blue}{One small hand
closed on the letter beside him and he slept on not knowing he was
special not knowing he was famous not knowing he would be woken in
a few hours’ time by Mrs Dursley’s scream as she opened the front
door to put out the milk bottles nor that he would spend the next
few weeks being prodded and pinched by his cousin Dudley.}
\textcolor{green}{Harry didn’t sleep all night. Perhaps it was
because he was now so busy what with Quidditch practice three
evenings a week on top of all his homework but Harry could hardly
believe it when he realized that he’d already been at Hogwarts two
months. Don’ mention it said Hagrid gruffly.}
\textcolor{blue}{Hagrid
grinned at Harry. I was allowed ter do a bit ter follow yeh an’ get
yer letters to yeh an’ stuff. There was only one room inside. }
\textcolor{red}{he leapt to his feet and ran to the window. }
\textcolor{orange}
{ It got to its feet and came swiftly toward
Harry. But he couldn’t do it. } 
\textcolor{blue}{He sat up and felt
around his eyes not used to the gloom.} 
\textcolor{red}{But he never wanted you dead.} 
\textcolor{green}{Hermione had now started
making study schedules for Harry and Ron too.}
\textcolor{red}{The
Chasers throw the Quaffle and put it through the hoops to score
Harry recited}

\subsection{Larger Versions of a Figure}
\label{app:sse:larger_fig}

A larger version of the clustering of digits shown in
Figure~\ref{fig:mnist} is shown on the next page. 

\clearpage

\begin{landscape}
\begin{figure}[t]
\hspace*{-1in}
\includegraphics[scale=0.37]{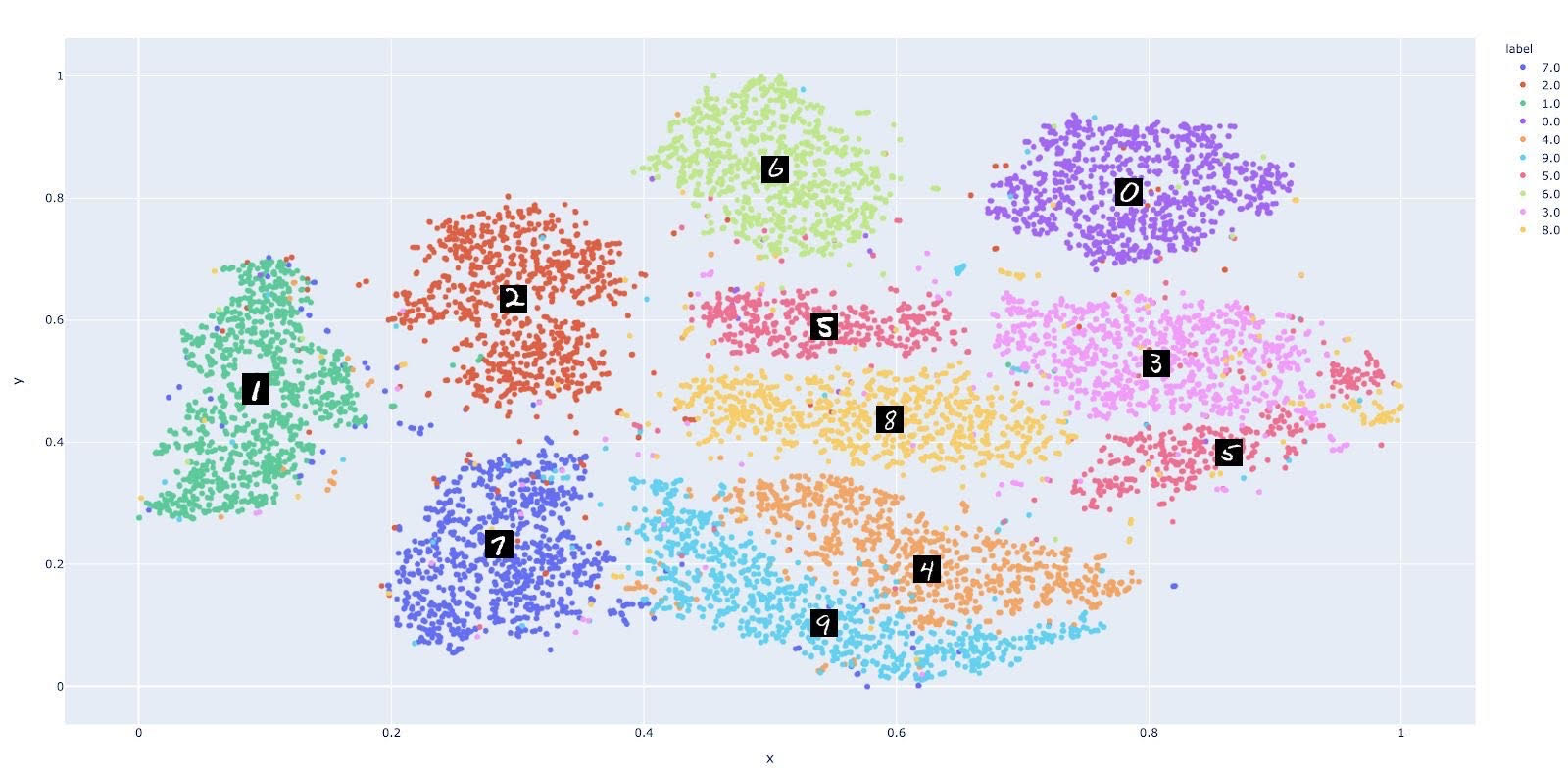}
\caption{Clusters and centroids (not exemplars) found by our 
method when applied to the MNIST dataset.}
\label{fig:mnist_large}
\end{figure}
\end{landscape}

\end{document}